\PassOptionsToPackage{table}{xcolor}
\documentclass[sigconf]{acmart}

\ifdefined\Bbbk\else\RequirePackage{amssymb}\fi

\usepackage{booktabs}
\usepackage{multirow}
\usepackage{amsmath}
\usepackage{caption}
\usepackage{xcolor}
\usepackage{balance}
\AtBeginDocument{%
  }

\copyrightyear{2026}
\acmYear{2026}
\setcopyright{cc}
\setcctype{by}

\acmConference[MM '26]
{Proceedings of the 34th ACM International Conference on Multimedia}
{November 10--14, 2026}
{Rio de Janeiro, Brazil}

\acmBooktitle{Proceedings of the 34th ACM International Conference on Multimedia
(MM '26), November 10--14, 2026, Rio de Janeiro, Brazil}

\acmDOI{10.1145/3767308.3835492}
\acmISBN{979-8-4007-2213-4/2026/11}

\title{SPOD: Recovering Suppressed Temporal Evidence in Video-LLMs via Single-Pass Orthogonal Decomposition}

\author{Yangguang Lin}
\email{sunshinelin@bupt.edu.cn}
\affiliation{%
  \institution{Beijing University of Posts and Telecommunications}
  \city{Beijing}
  \country{China}
}

\author{Quan Fang}
\authornote{Corresponding author.}
\email{qfang@bupt.edu.cn}
\affiliation{%
  \institution{Beijing University of Posts and Telecommunications}
  \city{Beijing}
  \country{China}
}

\author{Xiaoyan Liu}
\email{xiaoyanliu@bupt.edu.cn}
\affiliation{%
  \institution{Beijing University of Posts and Telecommunications}
  \city{Beijing}
  \country{China}
}

\author{Zefan Xu}
\email{xuzefan@bupt.edu.cn}
\affiliation{%
  \institution{Beijing University of Posts and Telecommunications}
  \city{Beijing}
  \country{China}
}

\renewcommand{\shortauthors}{Yangguang Lin, Quan Fang, Xiaoyan Liu, \& Zefan Xu}

\begin{document}

\begin{abstract}

Video Large Language Models (Video-LLMs) have achieved strong performance on many video understanding tasks, yet remain vulnerable to temporal hallucination in time-sensitive reasoning scenarios such as scene transition understanding and event ordering. Existing solutions either require costly retraining or introduce inefficient multi-pass inference and external supervision at test time. In this work, we show that temporal evidence is already encoded in Video-LLM hidden representations but is not fully exploited during standard inference. Motivated by this observation, we propose Single-Pass Orthogonal Decomposition (SPOD), a training-free framework that recovers suppressed temporal evidence through analytical hidden-state manipulation in a single forward pass. SPOD consists of two components: Frame Distinctiveness Analysis (FDA), which identifies salient frames from middle-layer hidden-state geometry, and Temporal Delta Subspace (TDS), which constructs orthogonal temporal and prior subspaces from weighted inter-frame hidden-state differences. By decomposing final hidden states into temporal, prior, and residual components and rescaling them independently, SPOD enhances dynamic information, suppresses static appearance bias, and preserves complementary information. Experiments on VidHalluc, EventHallusion, and VideoMMMU show that SPOD consistently outperforms strong inference-time baselines across temporal reasoning tasks and model architectures while retaining near-vanilla inference efficiency. These results suggest that improving temporal reasoning in Video-LLMs may depend not only on additional supervision, but also on recovering temporal evidence already present in their internal representations. Our code is available at \url{https://github.com/VioAgnes/SPOD}.

\end{abstract}


\maketitle

\section{Introduction}

Video Large Language Models (Video-LLMs) have advanced rapidly ~\cite{Lin2023Video-llava,Cheng2024VideoLLaMA2,Li2025LLaVA-OneVision,Li2025VideoChat-R1}, achieving performance in tasks such as video question answering~\cite{Fu2025Video-MME} and complex action recognition~\cite{Wang2023InternVideo}. Recent efforts like Video-R1~\cite{Feng2025Video-R1} further explore reinforcement learning to enhance video reasoning~\cite{Guo2025DeepSeek-R1}. Despite these advances, Video-LLMs still face challenges in temporal reasoning, frequently exhibiting \textbf{temporal hallucination}, where generated content contradicts the temporal logic of videos~\cite{Li2025VidHalluc,Zhang2024EventHallusion,Sun2024VideoHallucer}. This undermines their reliability in time-sensitive applications.

Existing approaches to temporal hallucination mainly follow two directions. Training-based methods~\cite{Li2025LLaVA-OneVision,Lin2023Video-llava,Bae2025MASH-VLM,Ding2025PaMi-VDPO} improve temporal modeling through retraining, but typically require large-scale data and substantial computational resources. Training-free methods avoid retraining, yet most either rely on contrastive decoding with multiple forward passes~\cite{Leng2024VisualContrastive,Ham2024mcd,Qi2025VideoCD} or depend on auxiliary calibration data and external vision models for activation manipulation~\cite{Sun2024TemporalInsight,Wu2025SEASON}. As a result, existing solutions often sacrifice inference efficiency or deployment simplicity. What remains missing is a practical training-free method that improves temporal reasoning without retraining, auxiliary supervision, or multi-pass decoding.

Rather than assuming that Video-LLMs fail because they lack temporal information~\cite{Li2025VidHalluc}, we ask whether temporal evidence is encoded in their hidden representations but underutilized during inference. To answer this question, we conduct an analysis of hidden states across temporal reasoning tasks. Our study reveals three findings (see Sec. \ref{sec:Preliminaries}). First, middle-layer hidden states exhibit frame discriminability that correlates with temporal saliency. Frames associated with scene changes or temporal moments produce distinctive geometric relationships with other frames in representation space. Second, frames selected by this discriminative structure capture meaningful and diverse temporal boundaries, rather than redundant visual variation. Third, inter-frame hidden-state differences lie in a low-dimensional subspace, indicating that temporal dynamics are not scattered in representation space, but are organized in a form amenable to explicit manipulation. Taken together, these observations suggest that the bottleneck is not the absence of temporal information, but such information is \textbf{systematically suppressed during standard inference}. 

\begin{figure}
  \centering
  \includegraphics[width=0.42\textwidth]{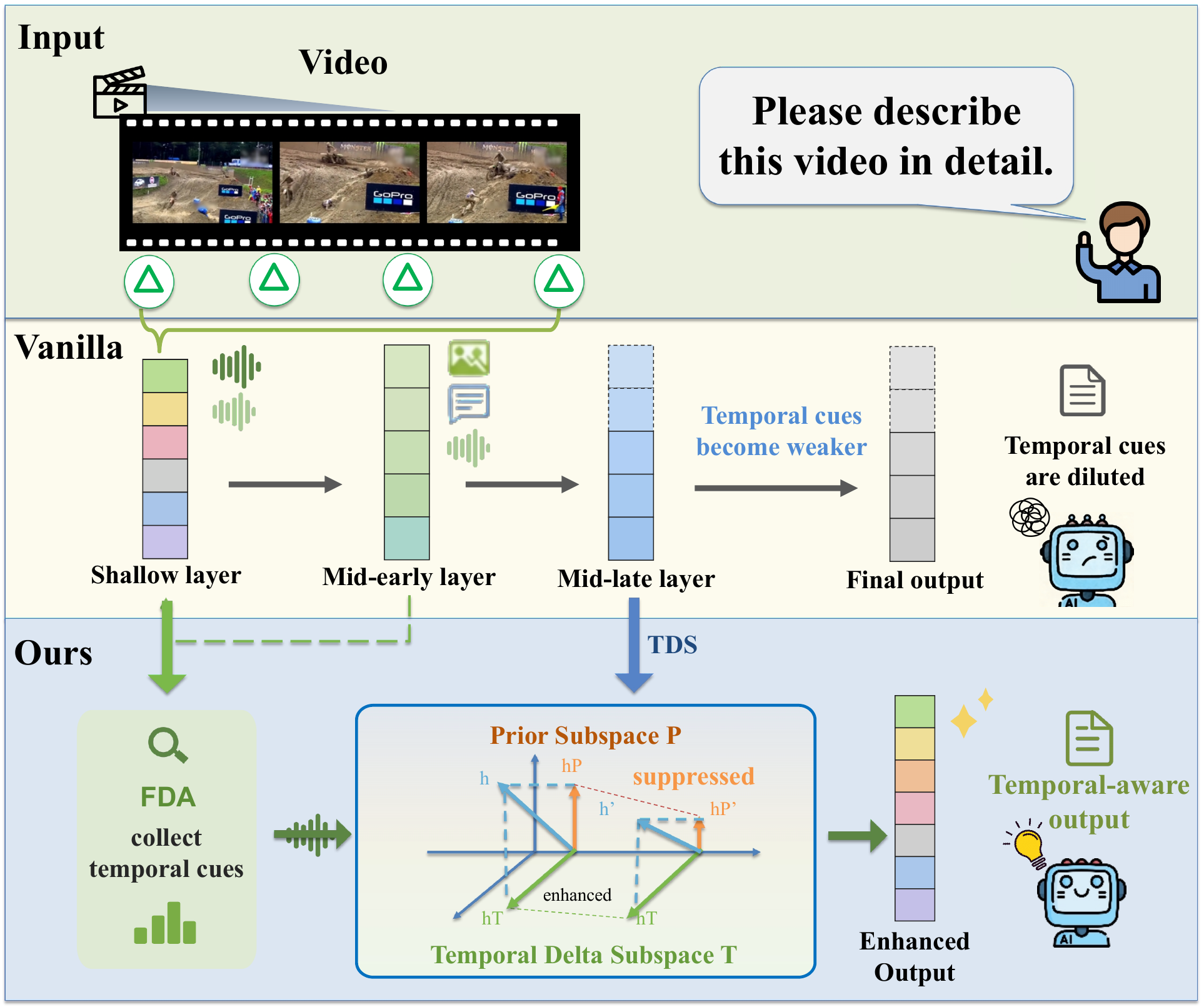}
  \caption{Illustration of temporal hallucination in Video-LLMs and our proposed SPOD.}
  \label{fig:teaser}
\end{figure}

Motivated by this observation, we propose \textbf{Single-Pass Orthogonal Decomposition (SPOD)}, a training-free framework that recovers temporal evidence through analytical hidden-state manipulation in a single forward pass. As illustrated in Figure~\ref{fig:teaser}, SPOD includes two components. \textbf{Frame Distinctiveness Analysis (FDA)} identifies temporally informative frames from middle-layer hidden-state geometry. \textbf{Temporal Delta Subspace (TDS)} constructs orthogonal temporal and prior subspaces from inter-frame hidden-state differences. SPOD then projects final-layer hidden states onto these subspaces and rescales them independently to enhance temporal information while suppressing static bias. This closed-form intervention improves temporal reasoning without retraining, iterative decoding, or external supervision.

Our main contributions are summarized as follows:
\begin{itemize} 
    \item \textbf{A structured analysis of temporal evidence in Video-LLMs.} Through systematic probing, we discover that temporal evidence is often already present in hidden representations but is not fully exploited during standard inference. 
    
    \item \textbf{A training-free framework for temporal evidence recovery.} We propose SPOD, a single-pass intervention that combines FDA and TDS to analytically recover temporal evidence, requiring no retraining, auxiliary supervision, external vision models, or iterative decoding.
    
    \item \textbf{Extensive validation across benchmarks and architectures.} Experiments on VidHalluc, EventHallusion, and VideoMMMU show that SPOD consistently outperforms strong training-free baselines across Video-LLM architectures and model scales, while maintaining near-vanilla inference efficiency.
    
\end{itemize}

\section{Related Work}

\subsection{Video Large Language Models}

Video-LLMs provide a unified paradigm for video understanding. Their architecture typically integrates a visual encoder, a projection layer that maps visual features into the language model's embedding space, and a large language model for text generation~\cite{Wu2024}. Video-LLaVA pioneered an alignment-before-projection approach, jointly learning visual and video representations through unified instruction tuning~\cite{Lin2023Video-llava}. Subsequent models such as LLaVA-OneVision and VideoChat-R1 advanced this paradigm through improved architectures and training strategies~\cite{Li2025LLaVA-OneVision,Li2025VideoChat-R1}. Recent efforts, including Video-R1, TinyLLaVA-Video-R1, and LongVLM, have explored reinforcement learning, test-time scaling, and long-video understanding~\cite{Feng2025Video-R1,Zhang2025TinyLLaVA-Video,Yu2024LongVideo,Guo2025DeepSeek-R1}. Efficient video token representations have also been studied to reduce computational overhead~\cite{Pan2024VideoToken,Wang2024SparseVideo}. Despite these advances, Video-LLMs still struggle with temporal reasoning due to the complexity of sequential visual information.

\subsection{Hallucinations in Video-LLMs}

Temporal hallucination is a critical challenge in Video-LLMs~\cite{Huang2024VideoHallucinationDetection,Zhao2024HallucinationSurvey}, where generated content contradicts video logic and undermines reliability. Several benchmarks have been developed for systematic evaluation. VidHalluc assesses QA and ranking tasks~\cite{Li2025VidHalluc}, EventHallusion evaluates event-level hallucinations~\cite{Zhang2024EventHallusion}, and VERHallu extends evaluation to causal relations~\cite{Zhang2026VERHallu}. Argus and Video-MMMU provide complementary evaluations of temporal and multimodal reasoning capabilities~\cite{Rawal2025Argus,Hu2025Video-MMMU}.

Existing mitigation methods fall into two categories. \textbf{Training-based methods} improve temporal modeling through caption supervision, attention refinement, preference optimization, or multimodal orchestration~\cite{Yang2024Vript,Bae2025MASH-VLM,Ding2025PaMi-VDPO,Chiu2025DaMO}, but require additional data and computation. \textbf{Training-free methods} intervene during inference. Contrastive decoding methods construct negatives from perturbed visual inputs to suppress language priors or enhance temporal reasoning~\cite{Leng2024VisualContrastive,Qi2025VideoCD,Wu2025SEASON}. Activation-based methods directly steer hidden states~\cite{Cai2025TAAE}; related image-based work also applies orthogonal subspace editing to mitigate object hallucination~\cite{Lin2026HulluEdit}. Other video-oriented methods explore visual-token reweighting, temporal grounding, and temporal focusing~\cite{Oquab2023DINOv2,Zheng2024TrainingFreeTG,Qian2024Momentor,Ding2026VideoZoomer}. In contrast, our work derives temporal subspaces from inter-frame hidden-state differences to mitigate video temporal hallucination without architectural modifications or contrastive decoding.

\subsection{Interpretability of Video-LLMs}

Recent interpretability studies examine how Video-LLMs process temporal information. Map the Flow shows that temporal reasoning emerges through cross-frame interactions in early-to-middle layers, followed by video-language alignment and response finalization in deeper layers~\cite{Kim2026MapFlow}. Complementary studies indicate that temporal information is structured across model depths and that temporal reasoning limitations may arise from the LLM rather than the visual encoder~\cite{Liu2025T3,Zhang2024ProbingLVLM,Zhang2025ProbingVideoTemporal}. LLMs may also possess latent temporal and causal reasoning capabilities recoverable from internal representations~\cite{Ko2023Temporal,Xiong2024Temporal}.

These findings motivate architectural and representation-level enhancement strategies. Architectural approaches such as Time Gating and TemporalAware modify model structure to improve temporal modeling~\cite{Hu2024TimeGating}. Layer-based methods exploit information across model depths. DoLa contrasts predictions from different layers to improve factuality~\cite{Chuang2023DoLa}, while Inference-Time Intervention steers hidden states to elicit more truthful responses~\cite{Li2023Inference-Time}. Although Temporal-Aware Activation Engineering also manipulates internal activations~\cite{Cai2025TAAE}, SPOD uses orthogonal subspace decomposition in a single forward pass without offline calibration or auxiliary datasets.

\section{Preliminary Analysis}
\label{sec:Preliminaries}

To investigate whether Video-LLMs encode temporal information internally that is underutilized during inference, we conduct three systematic analyses of hidden representations. To ensure that our observations generalize beyond a single model-video combination, we conduct analyses across three Video-LLMs using samples from temporal reasoning benchmarks. Frames are labeled as high- or low-saliency based on ground-truth scene-change annotations provided by these benchmarks, enabling a controlled comparison of hidden-state geometry at temporal versus static moments.

\begin{figure}[h!]
    \centering
    \includegraphics[width=0.42\textwidth]{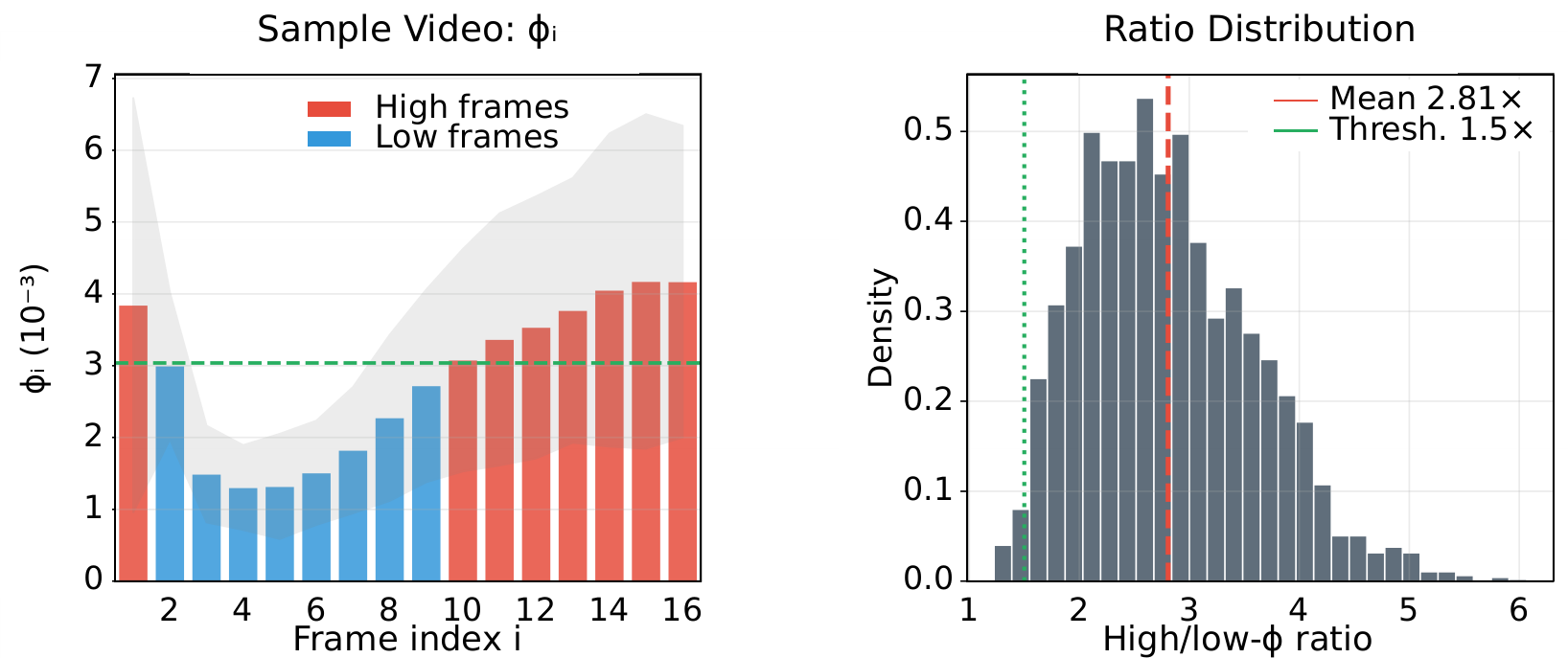}
    \caption{Frame-level temporal distinctiveness across video samples. Layer-aggregated scores \(\phi_i\) separate high- and low-saliency frames, with a mean high-to-low ratio of \(2.81\times\).}
    \label{fig:sigma_distribution}
\end{figure}

\textbf{Observation 1: Middle-layer hidden states encode inherent frame discriminability.} Our first analysis examines whether middle-layer hidden states encode frame-level discriminability. We measure frame distinctiveness $\phi_i$ as the layer-wise variance of pairwise cosine similarities between a frame and all others in the video (detailed in Sec. \ref{sec:frame_distinctiveness}). As illustrated in Figure \ref{fig:sigma_distribution}, the frame-wise similarity variance reveals a sharp distinction between static and dynamic content: the ratio exceeds a $1.5\times$ threshold in 98.7\% of cases, with a mean ratio of approximately $2.8\times$, and the signal peaks near layer 10. These findings suggest that middle-layer hidden states are sensitive to frame-level distinctiveness in ways that correlate with temporal saliency, providing empirical grounding for investigating temporal evidence recovery in subsequent layers.

\begin{figure}[h!]
    \centering
    \includegraphics[width=0.42\textwidth]{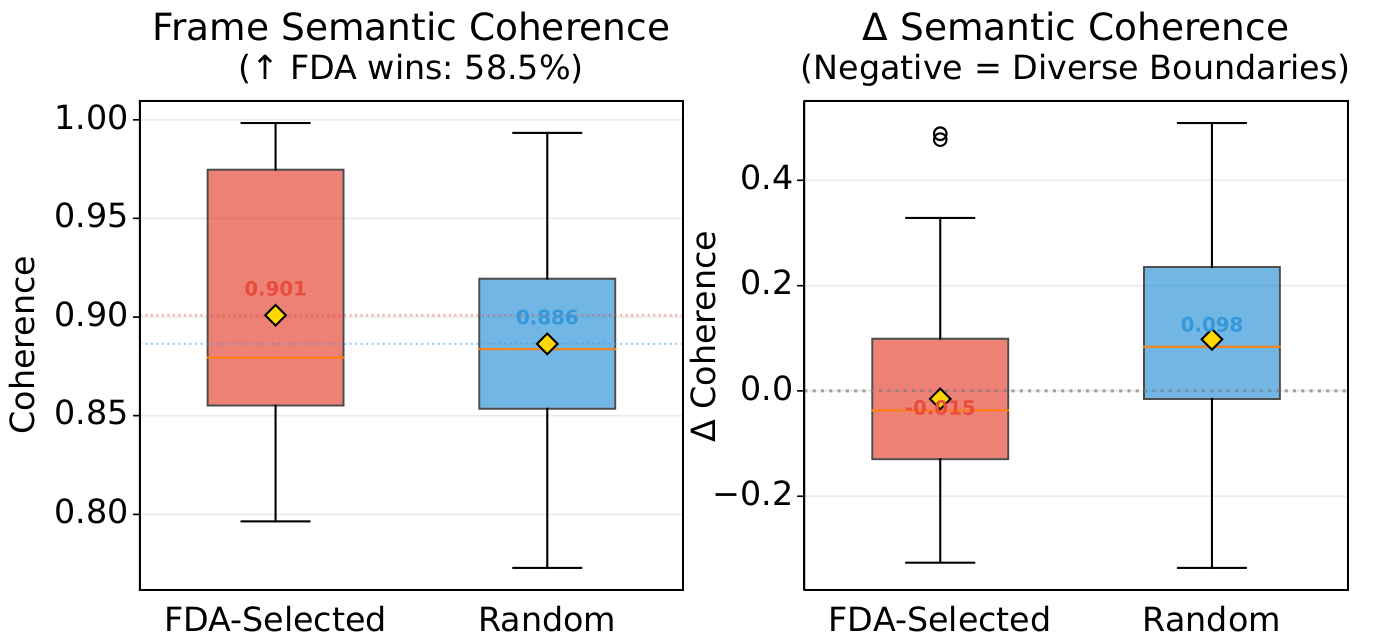}
    \caption{Comparison of frame semantic coherence and temporal-boundary diversity between FDA-selected and uniformly sampled random frames.}
    \label{fig:fda_vs_random}
\end{figure}

\textbf{Observation 2: Frames identified by discriminability analysis capture diverse temporal boundaries.} Having established that discriminative signals exist, we next examine whether these signals capture semantically meaningful temporal boundaries. To assess the quality of these signals, we compare frames selected by FDA against uniformly random frames, as shown in Fig.~\ref{fig:fda_vs_random}. FDA-selected frames achieve superior semantic consistency while simultaneously exhibiting more diverse temporal boundaries: the mean frame semantic coherence is $0.901$ versus $0.886$ for random sampling, with FDA winning in 58.5\% of samples. The frequent occurrence of negative $\Delta$ consistency variations (mean $-0.015$ versus $0.098$) indicates lower directional redundancy among the captured changes, suggesting that the selected frames represent distinct events rather than redundant information. Our findings confirm that the model has the ability to distinguish semantically rich temporal boundaries through its latent representations.

\begin{figure}[h!]
    \centering
    \includegraphics[width=0.42\textwidth]{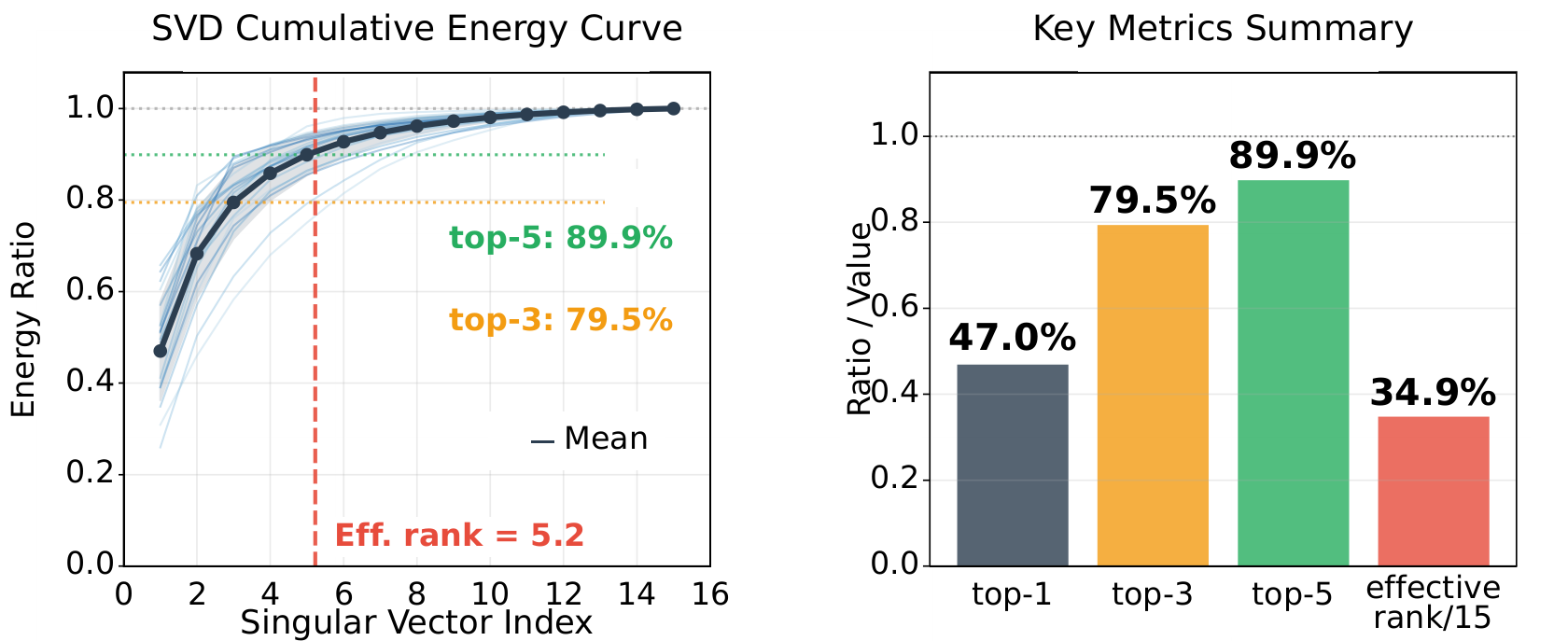}
    \caption{Cumulative singular-value energy and effective-rank statistics of FDA-weighted temporal-difference matrices used for temporal subspace construction.}
    \label{fig:tds_svd}
\end{figure}

\textbf{Observation 3: The temporal difference subspace is compact and structurally organized.} Finally, we investigate whether inter-frame differences form a structured subspace suitable for manipulation. We perform singular value decomposition on inter-frame difference vectors to characterize the geometry of the temporal subspace. As shown in Fig.~\ref{fig:tds_svd}, the top three singular vectors account for 79.5\% of the cumulative energy, while the first five account for approximately 90\%, with an effective rank of 5.2. These results suggest that temporal variations are concentrated in a low-dimensional subspace. Moreover, the low directional consistency between different difference vectors suggests that distinct temporal events are encoded along diverse, weakly aligned directions in hidden space. \textbf{These results support modeling temporal evidence as a compact and manipulable subspace.}

These empirical findings motivate our proposed approach. Building on the observation that temporal evidence already exists within hidden representations yet remains suppressed during standard inference, we develop a method that explicitly extracts and enhances these signals through orthogonal subspace decomposition, thereby enabling more effective temporal evidence recovery without requiring additional training or external signals.

\begin{figure*}[t]
    \centering
    \includegraphics[width=\textwidth]{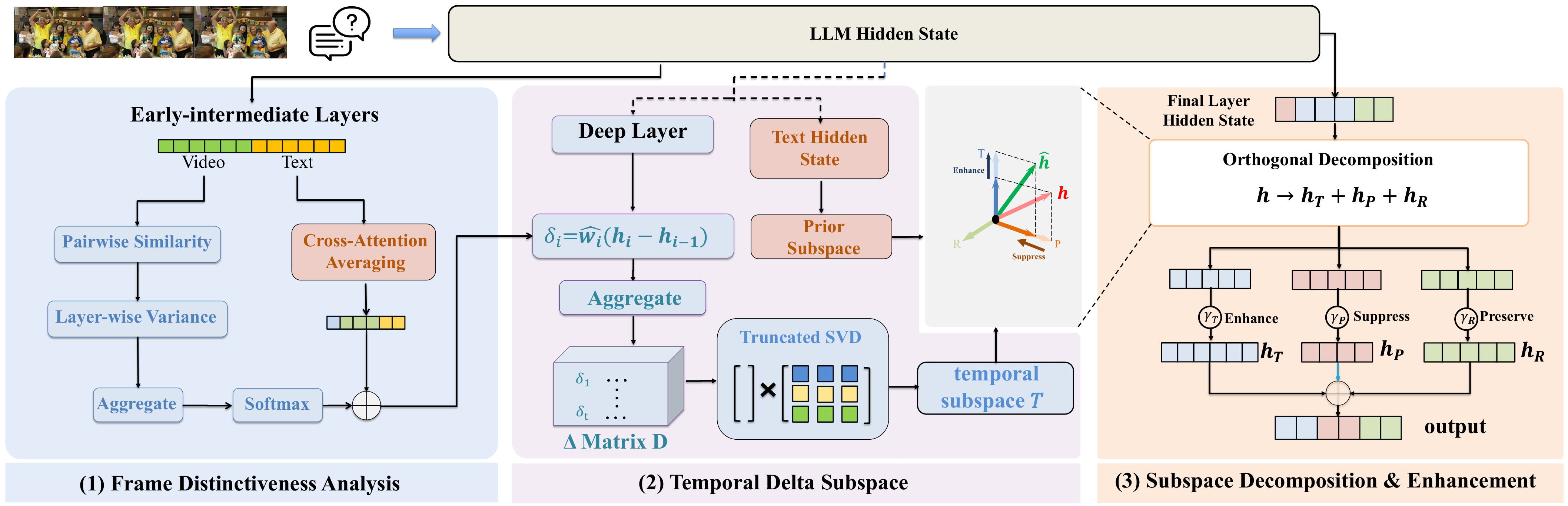}
    \caption{SPOD pipeline. (1) FDA identifies salient frames from middle-layer hidden-state geometry. (2) TDS constructs temporal and prior subspaces from weighted inter-frame differences and text hidden states. (3) Final hidden states are decomposed into temporal, prior, and residual components and rescaled to recover temporal evidence in a single forward pass.}
    \label{fig:framework}
\end{figure*}

\section{Methodology}

\subsection{Overview of SPOD}

As illustrated in Fig.~\ref{fig:framework}, SPOD recovers suppressed temporal evidence in a single inference pass. FDA identifies temporally salient frames, while TDS constructs temporal and prior subspaces from weighted inter-frame differences and text hidden states. Component-wise rescaling then enhances temporal evidence while adaptively adjusting prior and preserving residual components.

\subsection{Frame Distinctiveness Analysis}
\label{sec:frame_distinctiveness}
FDA aims to identify temporally salient frames by exploiting an inherent property of Video-LLMs: in middle layers, frames carrying significant temporal information produce hidden states that are structurally distinct from those of neighboring frames. Unlike prior approaches that rely on external cues such as pixel differences, saliency maps from frozen vision models, or dual-rate processing, FDA extracts the discriminative signal directly from the LLM's internal representations, following the spirit of DoLa and ITI.

Let $\mathbf{H}_i^{(l)} \in \mathbb{R}^d$ denote the mean hidden state of frame $i$ at layer $l$, and let $\mathcal{L} = \{l_1, \dots, l_m\}$ denote the FDA layer range.

\textbf{Pairwise Similarity Computation.} For each frame $i$ and layer $l \in \mathcal{L}$, we normalize hidden states to unit norm and compute the full pairwise cosine similarity matrix:
\begin{equation}
\tilde{\mathbf{H}}_i^{(l)} = \frac{\mathbf{H}_i^{(l)}}{\|\mathbf{H}_i^{(l)}\|}, \quad
\mathbf{S}^{(l)} = \tilde{\mathbf{H}}^{(l)} \tilde{\mathbf{H}}^{(l)\top}
\end{equation}
where $\mathbf{S}^{(l)} \in \mathbb{R}^{T \times T}$ with entries $s_{ij}^{(l)} = \tilde{\mathbf{H}}_i^{(l)\top} \tilde{\mathbf{H}}_j^{(l)}$ and $s_{ii}^{(l)} = 1$.

\textbf{Layer-wise Variance.} The variance of similarities in row $i$ of $\mathbf{S}^{(l)}$ quantifies how unevenly frame $i$ relates to others:
\begin{equation}
\phi_i^{(l)} = \mathrm{Var}_j(s_{ij}^{(l)}) = \frac{1}{T} \sum_{j=1}^{T} \left(s_{ij}^{(l)} - \bar{s}_i^{(l)}\right)^2
\end{equation}
where $\bar{s}_i^{(l)} = \frac{1}{T} \sum_{j=1}^{T} s_{ij}^{(l)}$ is the sample mean similarity and $T$ is the total number of frames. Here we adopt the maximum likelihood estimator for variance, which is consistent with the mean estimation above. High variance indicates that frame $i$ is strongly similar to some frames and strongly dissimilar to others, characteristic of frames marking scene transitions or salient temporal moments.

\textbf{Layer Aggregation and Smoothing.} We aggregate across $\mathcal{L}$ and apply a smoothing window of size 3:
\begin{equation}
\varphi_i = \mathrm{Smooth}_3\!\left(\frac{1}{|\mathcal{L}|} \sum_{l \in \mathcal{L}} \phi_i^{(l)}\right)
\label{eq:smooth_phi}
\end{equation}
where $\varphi_i$ is the smoothed, layer-averaged distinctiveness score for frame $i$.

\textbf{Softmax Normalization.} Scores are normalized to obtain frame weights:
\begin{equation}
w_i = \frac{\exp(\varphi_i)}{\sum_{j=1}^{T} \exp(\varphi_j)}
\label{eq:frame_weights}
\end{equation}
where $w_i$ reflects the model's hidden-state-level assessment of temporal evidence.

\textbf{Attention-Level Refinement.} To further align FDA with the LLM's cross-modal reasoning pathway, we refine the weights using cross-attention from text tokens to video tokens. We average cross-attention weights across heads and question tokens at each layer, then aggregate across $\mathcal{L}$ to obtain a per-frame textual relevance score $a_i$. The final frame weight is a convex combination:
\begin{equation}
\hat{w}_i = \alpha \cdot w_i + (1-\alpha) \cdot a_i
\label{eq:refined_weights}
\end{equation}
where $\alpha \in [0, 1]$ balances hidden-state distinctiveness and textual relevance. These refined weights $\hat{w}_i$ guide the construction of the Temporal Delta Subspace.
\subsection{Temporal Delta Subspace}
\label{sec:temporal_delta_subspace}
TDS constructs an orthogonal subspace that explicitly captures temporal dynamics, complementing FDA's frame identification. A key limitation of existing subspace-based approaches is that they apply PCA or SVD directly to video token sets, treating frames as unordered points in representation space and ignoring temporal order. In contrast, TDS builds the subspace from \textbf{inter-frame hidden-state differences}, which naturally encode temporal change. We extract these representations from a deep layer $l^*$ near the Stage 2/3 transition boundary.

\textbf{Temporal Delta Construction.} We extract hidden states at layer $l^*$ and compute weighted inter-frame differences:
\begin{equation}
\delta_i = \hat{w}_i \cdot \left(\mathbf{H}_i^{(l^*)} - \mathbf{H}_{i-1}^{(l^*)}\right), \quad \forall i \in \{2, \dots, T\}
\label{eq:delta_vectors}
\end{equation}
where $\delta_i$ encodes the direction and magnitude of representational change from frame $i-1$ to frame $i$, weighted by its temporal significance from FDA.

\textbf{Truncated SVD for Temporal Subspace.} We stack these differences into a matrix $\Delta \in \mathbb{R}^{(T-1) \times d}$ and apply truncated SVD:
\begin{equation}
\Delta = \mathbf{U} \boldsymbol{\Sigma} \mathbf{V}^\top
\label{eq:delta_svd}
\end{equation}
where $\mathbf{U} \in \mathbb{R}^{(T-1) \times (T-1)}$ is a square orthogonal matrix, $\boldsymbol{\Sigma} \in \mathbb{R}^{(T-1) \times d}$ is a rectangular matrix containing the singular values $\sigma_1 \geq \sigma_2 \geq \cdots \geq \sigma_{T-1} \geq 0$ on its first $(T-1)$ diagonal entries (with remaining entries being zero), and $\mathbf{V} \in \mathbb{R}^{d \times d}$ is an orthogonal matrix of right singular vectors. We then take $\mathbf{V}_r = \mathbf{V}^{(1:r)} \in \mathbb{R}^{d \times r}$, the top-$r$ right singular vectors forming an orthonormal basis for the temporal subspace $\mathcal{T}$. By building from inter-frame differences rather than absolute token positions, this subspace directly encodes the directions of temporal change; frames near scene transitions produce large magnitudes in these difference vectors, and their directions dominate the top singular vectors.

\textbf{Prior Subspace Construction.} To construct the prior subspace $\mathcal{P}$, we project text token hidden states onto the orthogonal complement of $\mathcal{T}$. Let $\mathbf{T} \in \mathbb{R}^{n_t \times d}$ denote the matrix of $n_t$ text token hidden states at layer $l^*$. We compute:
\begin{equation}
\mathbf{T}_{\perp} = \mathbf{T} \left(\mathbf{I}_d - \mathbf{V}_r \mathbf{V}_r^\top\right) \in \mathbb{R}^{n_t \times d}
\label{eq:text_proj}
\end{equation}
which removes the temporally aligned component from each text token. We then perform SVD:
\begin{equation}
\mathbf{T}_{\perp} = \mathbf{U}' \Sigma' \mathbf{V}'^\top \quad \Rightarrow \quad \mathbf{P}_q = \mathbf{V}'^{(1:q)} \in \mathbb{R}^{d \times q}
\label{eq:prior_svd}
\end{equation}
where $\mathbf{P}_q = \mathbf{V}'^{(1:q)}$ consists of the first $q$ columns of $\mathbf{V}'$, forming the orthonormal basis for the prior subspace $\mathcal{P}$. By construction, $\mathbf{V}_r^\top \mathbf{P}_q = \mathbf{0}$, confirming $\mathcal{P} \perp \mathcal{T}$.

\textbf{Semantic Interpretation.} The temporal subspace $\mathcal{T} = \mathrm{span}(\mathbf{V}_r)$ encodes temporal dynamics: directions of change between consecutive frames, including scene transitions, motion patterns, and event ordering. The prior subspace $\mathcal P$ mainly captures static, non-temporal visual-semantic content, such as object identity, spatial layout, and scene context. The residual subspace spans directions orthogonal to both $\mathcal{T}$ and $\mathcal{P}$.

For temporal reasoning tasks, the answer is fundamentally determined by \emph{what changes} rather than \emph{what remains static}. Empirically, scaling down the static component while scaling up the dynamic component improves accuracy. We do not claim $\mathcal{P}$ exclusively captures task-irrelevant content; static appearance can be essential for non-temporal questions.

\subsection{Subspace Decomposition and Enhancement}
\label{sec:subspace_decomposition}
With the temporal subspace $\mathcal{T}$ and prior subspace $\mathcal{P}$ established, we describe the decomposition and enhancement mechanism for recovering suppressed temporal evidence in a single forward pass.

\textbf{Decomposition.} Given a hidden state $\mathbf{h} \in \mathbb{R}^d$ at the final layer, we decompose it via orthogonal projections onto the three mutually orthogonal subspaces:
\begin{align}
\mathbf{h}_\mathcal{T} &= \mathbf{V}_r \mathbf{V}_r^\top \mathbf{h} \label{eq:decomp_temporal} \\
\mathbf{h}_\mathcal{P} &= \mathbf{P}_q \mathbf{P}_q^\top \mathbf{h} \label{eq:decomp_prior} \\
\mathbf{h}_\mathcal{R} &= \left(\mathbf{I}_d - \mathbf{V}_r \mathbf{V}_r^\top - \mathbf{P}_q \mathbf{P}_q^\top\right) \mathbf{h} \label{eq:decomp_residual}
\end{align}
where $\mathbf{h}_\mathcal{T}$, $\mathbf{h}_\mathcal{P}$, and $\mathbf{h}_\mathcal{R}$ are the temporal, prior, and residual components, respectively.

\textbf{Enhancement.} For enhanced temporal reasoning, we scale each component independently:
\begin{equation}
\hat{\mathbf{h}} = \gamma_\mathcal{T} \cdot \mathbf{h}_\mathcal{T} + \gamma_\mathcal{P} \cdot \mathbf{h}_\mathcal{P} + \gamma_\mathcal{R} \cdot \mathbf{h}_\mathcal{R}
\label{eq:enhancement}
\end{equation}
where $\gamma_\mathcal{T} \geq 1$ enhances temporal evidence, $\gamma_\mathcal{P} \in (0, 1)$ suppresses prior components, and $\gamma_\mathcal{R} = 1$ preserves the residual unchanged. Since the bases are orthonormal and mutually orthogonal, the components can be manipulated independently without cross-interference. This enables temporal evidence recovery in a single forward pass, without iterative optimization or parameter updates.

\subsection{Complexity Analysis}
For a video with $T$ frames and hidden dimension $d$, FDA computes pairwise cosine similarities across the selected layers $\mathcal{L}$, yielding a complexity of $\mathcal{O}(T^2d|\mathcal{L}|)$, where the normalization cost $\mathcal{O}(Td|\mathcal{L}|)$ is dominated. For TDS, the temporal-difference matrix $\Delta \in \mathbb{R}^{(T-1)\times d}$ satisfies $T-1<d$, as typically $d \gg T$ in Video-LLMs. Applying thin SVD to $\Delta$ costs $\mathcal{O}((T-1)^2d)=\mathcal{O}(T^2d)$. In the final decomposition step, projecting each generated hidden state onto the temporal and prior subspaces using low-rank bases $\mathbf{V}_r \in \mathbb{R}^{d\times r}$ and $\mathbf{P}_q \in \mathbb{R}^{d\times q}$ costs $\mathcal{O}(dr+dq)$ per step, while rescaling is negligible. Therefore, the extra overhead is dominated by FDA similarity computation and SVD, and SPOD remains lightweight compared with a full model forward pass while requiring no iterative optimization.

\section{Experiments}
\label{sec:experiments}

\subsection{Experimental Settings}

\textbf{Dataset.} We evaluate SPOD on three main benchmarks: VidHalluc~\cite{Li2025VidHalluc}, EventHallusion~\cite{Zhang2024EventHallusion}, and VideoMMMU~\cite{Hu2025Video-MMMU}. VidHalluc quantifies temporal hallucinations across four task categories including binary question answering, multiple-choice questions, scene-time hallucination, and temporal scene ranking. EventHallusion assesses event-level hallucinations through Entire, Mix, and Misleading subtasks. VideoMMMU provides a comprehensive evaluation spanning Perception, Comprehension, and Adaptation tasks. Our experiments on VidHalluc follow the official validation set protocol while EventHallusion results are reported on the full benchmark as distributed.

\textbf{Models.} We evaluate three 7B Video-LLMs: Qwen2.5-VL-7B~\cite{An2025Qwen25VL}, VideoLLaMA2-7B~\cite{Cheng2024VideoLLaMA2} and Video-LLaVA-7B~\cite{Lin2023Video-llava}. These cover diverse architectures, including dual-encoder and unified encoding strategies with varying backbones. All experiments uniformly sample 16 frames per video without any fine-tuning. To further validate the scalability of SPOD, we additionally conduct experiments on Qwen2.5-VL-72B, a state-of-the-art large-scale Video-LLM that represents a significant leap in model capacity.

\textbf{Baselines.} We compare against a vanilla model (no intervention) and five representative inference-time baselines. These baselines are categorized into three groups based on their mechanisms. The first group comprises contrastive decoding methods: TCD~\cite{Zhang2024EventHallusion} contrasts predictions from clean inputs against frame-dropped counterparts to identify spurious visual priors. The second group includes external-model-assisted methods: CGD~\cite{Leng2024VisualContrastive} and VTI~\cite{Liu2025Reducing} refine generation by leveraging external frozen vision models, where CGD employs contrastive logits and VTI performs temporal hidden-state steering through auxiliary vision guidance. The third group consists of auxiliary-training methods: DINO-HEAL~\cite{Li2025VidHalluc} reweights video tokens through DINOv2 saliency maps~\cite{Oquab2023DINOv2} using a separately trained saliency module, while TA-AE~\cite{Cai2025TAAE} reconstructs video features through temporal autoencoding with a dedicated auxiliary autoencoder. During evaluation, all baselines are applied without updating the underlying Video-LLM parameters.

\textbf{Implementation details.} For Frame Distinctiveness Analysis, we compute distinctiveness scores using hidden states from layers 8 to 16, refined via cross-attention weights from text to video tokens with fusion weight $\alpha=0.5$. For Temporal Delta Subspace, we extract inter-frame differences from layer 22 and perform truncated SVD to obtain an orthogonal basis. The temporal subspace rank $r$ is set to 4, prior subspace rank $q$ to 4, and scaling factors are configured as $\gamma_\mathcal{T}=1.2$, $\gamma_\mathcal{P}=0.5$, $\gamma_\mathcal{R}=1.0$. All experiments follow official evaluation protocols using deterministic generation with $\text{Temperature}=0$ and $\text{Top-}k=1$, and all baselines are reproduced according to their original papers. All experiments are conducted on NVIDIA A100 GPUs with 40GB memory. For VideoLLaMA2-7B and Video-LLaVA-7B, the FDA layer range and TDS extraction layer are adjusted proportionally to their total depth to preserve the same relative positions within their Stage 2 region.

\begin{table}[t]
    \centering
    \caption{Results on VidHalluc benchmark. \textbf{Bold} indicates the best result per metric within each model group.}
    \label{tab:vidhalluc_results}
    \begin{tabular}{lccccc}
        \toprule
        \textbf{Model} & \textbf{BQA} & \textbf{MCQ} & \textbf{STH} & \textbf{TSH} & \textbf{Overall} \\
        \midrule
        \rowcolor{gray!22}\multicolumn{6}{c}{\textit{VideoLLaMA2}} \\
        Vanilla & 75.77 & 83.35 & 56.55 & 58.17 & 68.46 \\
        TCD & 76.77 & 83.65 & 55.19 & 56.67 & 68.07 \\
        DINO-HEAL & 75.79 & 83.35 & 56.32 & 57.67 & 68.28 \\
        TA-AE & 79.09 & 84.07 & 65.14 & 61.67 & 72.49 \\
        CGD & 76.03 & 82.76 & 59.14 & 59.02 & 69.24 \\
        VTI & 78.14 & 83.67 & 62.47 & 60.28 & 71.14 \\
        Ours & \textbf{81.48} & \textbf{84.28} & \textbf{69.80} & \textbf{70.67} & \textbf{76.56} \\
        \midrule
        \rowcolor{gray!22}\multicolumn{6}{c}{\textit{Qwen2.5-VL}} \\
        Vanilla & 73.50 & 63.66 & 60.55 & 59.17 & 64.22 \\
        TCD & 75.49 & 74.35 & 46.81 & 65.83 & 65.62 \\
        DINO-HEAL & 69.77 & 65.69 & 60.97 & 59.83 & 64.07 \\
        TA-AE & 76.18 & 74.95 & 61.81 & 61.33 & 68.57 \\
        CGD & 74.23 & 65.18 & 60.97 & 59.86 & 65.06 \\
        VTI & 75.49 & 69.82 & 61.74 & 60.97 & 67.01 \\
        Ours & \textbf{82.16} & \textbf{76.28} & \textbf{67.40} & \textbf{68.22} & \textbf{73.52} \\
        \midrule
        \rowcolor{gray!22}\multicolumn{6}{c}{\textit{Video-LLaVA}} \\
        Vanilla & 67.75 & 66.60 & 21.80 & 46.83 & 50.75 \\
        TCD & 70.40 & 65.97 & 21.77 & 42.33 & 50.12 \\
        DINO-HEAL & 70.82 & 67.19 & 26.47 & 47.50 & 53.00 \\
        TA-AE & 67.70 & 66.84 & 41.16 & 49.50 & 56.30 \\
        CGD & 67.42 & 65.79 & 29.17 & 47.29 & 52.42 \\
        VTI & 68.23 & 66.14 & 33.24 & 48.96 & 54.14 \\
        Ours & \textbf{71.30} & \textbf{68.06} & \textbf{48.36} & \textbf{66.50} & \textbf{63.56} \\
        \bottomrule
    \end{tabular}
\end{table}

\subsection{Main Results on Hallucination Benchmarks}
\label{sec:main_results}

\textbf{VidHalluc Results.} We evaluate SPOD on VidHalluc, a benchmark specifically designed to assess temporal hallucination. As shown in Table~\ref{tab:vidhalluc_results}, SPOD delivers consistent and substantial improvements across all three backbone models, with an average overall gain of 10.07 points over the vanilla models. On VideoLLaMA2, SPOD improves the overall score from 68.46 to 76.56, with the largest gains in scene time hallucination, from 56.55 to 69.80, and temporal scene hallucination, from 58.17 to 70.67. On Qwen2.5-VL, SPOD achieves an overall gain of 9.3 points and improves every metric, most notably temporal scene hallucination from 59.17 to 68.22. Similar improvements are observed on Video-LLaVA, where the overall score increases from 50.75 to 63.56 and the temporal scene hallucination score rises substantially from 46.83 to 66.50. These results demonstrate that SPOD effectively mitigates temporal hallucination by recovering suppressed temporal evidence through orthogonal subspace decomposition, with particularly clear benefits for challenging scene level temporal relations.

\begin{table}[h]
    \centering
    \caption{Results on Qwen2.5-VL-72B.}
    \label{tab:large_scale_results}
    \begin{tabular}{l|cccc}
        \toprule
        \textbf{Variant} & \textbf{BQA} & \textbf{MCQ} & \textbf{STH} & \textbf{TSH} \\
        \midrule
        Vanilla & 81.47 & 81.24 & 67.60 & 84.00 \\
        Ours & \textbf{84.73} & \textbf{88.45} & \textbf{68.64} & \textbf{85.67} \\
        \bottomrule
    \end{tabular}
\end{table}

\textbf{Large-Scale Generalization.} To validate the scalability of SPOD, we extend our experiments to Qwen2.5-VL-72B, a state-of-the-art large-scale Video-LLM. As shown in Table~\ref{tab:large_scale_results}, our method maintains consistent improvements across all metrics. Notably, larger models already possess stronger temporal understanding capabilities and reduced hallucination rates, so the absolute gains are naturally smaller than on 7B models. The consistent relative gains nonetheless confirm that SPOD generalizes to larger architectures and effectively recovers temporal evidence at greater model scales.

\textbf{EventHallusion Results.} On EventHallusion, which evaluates event level reasoning through Entire, Mix, and Misleading subtasks, SPOD achieves the best overall score across all three backbone models. As shown in Table~\ref{tab:eventhallusion_results}, our method demonstrates particularly strong performance on the Misleading subtask, with VideoLLaMA2 improving from 46.08 to 86.27. On Video-LLaVA, SPOD increases the Misleading score from 41.18 to 91.00, demonstrating robust reasoning under misleading temporal content. After including the inference time baselines CGD and VTI, SPOD remains the strongest overall method across all evaluated backbones. Although individual baselines may perform better on particular subtasks, the consistent overall gains indicate that SPOD improves event level temporal reasoning without sacrificing performance across different question types. These results confirm that recovered temporal evidence generalizes from frame-level hallucination to event-level reasoning.

\begin{table}[t]
    \centering
    \caption{Results on EventHallusion benchmark.}
    \label{tab:eventhallusion_results}
    \begin{tabular}{lcccc}
        \toprule
        \textbf{Model} & \textbf{Entire} & \textbf{Mix} & \textbf{Misleading} & \textbf{Overall} \\
        \midrule
        \rowcolor{gray!22}\multicolumn{5}{c}{\textit{VideoLLaMA2}} \\
        \hspace{1em}Vanilla & 38.60 & 62.18 & 46.08 & 48.95 \\
        \hspace{1em}TCD & 42.98 & \textbf{75.65} & 54.90 & 57.84 \\
        \hspace{1em}DINO-HEAL & 38.60 & 62.69 & 46.08 & 48.96 \\
        \hspace{1em}TA-AE & 43.86 & 70.47 & 57.70 & 57.34 \\
        \hspace{1em}CGD & 42.11 & 63.21 & 47.06 & 50.79 \\
        \hspace{1em}VTI & 38.60 & 67.36 & 46.08 & 50.68 \\
        \hspace{1em}Ours & \textbf{46.49} & 71.00 & \textbf{86.27} & \textbf{67.92} \\
        \midrule
        \rowcolor{gray!22}\multicolumn{5}{c}{\textit{Qwen2.5-VL}} \\
        \hspace{1em}Vanilla & 71.93 & 39.38 & 98.04 & 69.78 \\
        \hspace{1em}TCD & 69.30 & 43.01 & 97.06 & 69.79 \\
        \hspace{1em}DINO-HEAL & 80.70 & 35.75 & 95.10 & 70.52 \\
        \hspace{1em}TA-AE & 89.47 & 41.50 & \textbf{100.00} & 76.99 \\
        \hspace{1em}CGD & 71.93 & 32.12 & 99.02 & 67.69 \\
        \hspace{1em}VTI & \textbf{99.12} & 27.98 & \textbf{100.00} & 75.70 \\
        \hspace{1em}Ours & 89.80 & \textbf{43.50} & \textbf{100.00} & \textbf{77.77} \\
        \midrule
        \rowcolor{gray!22}\multicolumn{5}{c}{\textit{Video-LLaVA}} \\
        \hspace{1em}Vanilla & 41.23 & 57.71 & 41.18 & 46.70 \\
        \hspace{1em}TCD & 46.49 & 54.92 & 74.50 & 58.64 \\
        \hspace{1em}DINO-HEAL & 39.47 & 48.71 & 79.41 & 55.86 \\
        \hspace{1em}TA-AE & 79.83 & 55.96 & 90.20 & 75.33 \\
        \hspace{1em}CGD & 40.40 & 47.70 & 66.70 & 51.60 \\
        \hspace{1em}VTI & 56.14 & \textbf{65.80} & 58.82 & 60.25 \\
        \hspace{1em}Ours & \textbf{81.60} & 63.20 & \textbf{91.00} & \textbf{78.60} \\
        \bottomrule
    \end{tabular}
\end{table}

\subsection{VideoMMMU Results}
\label{sec:videommmu}

We further evaluate SPOD on VideoMMMU, a comprehensive benchmark designed to assess general video understanding capabilities across perception, comprehension, and adaptation tasks. Unlike the hallucination-focused benchmarks, VideoMMMU tests whether enhanced temporal reasoning translates to broader video comprehension abilities. As shown in Table~\ref{tab:videommmu_results}, after adding TCD and TA-AE as inference-time baselines, SPOD achieves the best overall score across all three backbone models. On Qwen2.5-VL, the overall score improves from 47.44 for vanilla and 49.00 for TA-AE to 54.33. Similar overall gains are observed on VideoLLaMA2 and Video-LLaVA. Although some baselines perform better on individual subcategories, SPOD provides the most consistent overall improvement, suggesting broad applicability to real-world video comprehension tasks where temporal reasoning is essential.

\begin{table}[t]
    \centering
    \caption{Results on VideoMMMU benchmark.}
    \label{tab:videommmu_results}
    \begin{tabular}{lcccc}
        \toprule
        \textbf{Model} & \textbf{Perc.} & \textbf{Comp.} & \textbf{Adapt.} & \textbf{Overall} \\
        \midrule
        \rowcolor{gray!22}\multicolumn{5}{c}{\textit{VideoLLaMA2}} \\
        \hspace{1em}Vanilla & 60.34 & 46.00 & 27.67 & 44.67 \\
        \hspace{1em}TCD & 60.00 & 43.00 & \textbf{36.33} & 46.44 \\
        \hspace{1em}TA-AE & 61.67 & 43.00 & 35.00 & 46.56 \\
        \hspace{1em}Ours & \textbf{68.67} & \textbf{48.34} & 28.67 & \textbf{48.56} \\
        \midrule
        \rowcolor{gray!22}\multicolumn{5}{c}{\textit{Qwen2.5-VL}} \\
        \hspace{1em}Vanilla & 58.33 & 44.33 & 39.67 & 47.44 \\
        \hspace{1em}TCD & 62.00 & 45.00 & 29.00 & 45.33 \\
        \hspace{1em}TA-AE & 64.00 & \textbf{51.00} & 32.00 & 49.00 \\
        \hspace{1em}Ours & \textbf{68.33} & 50.33 & \textbf{44.33} & \textbf{54.33} \\
        \midrule
        \rowcolor{gray!22}\multicolumn{5}{c}{\textit{Video-LLaVA}} \\
        \hspace{1em}Vanilla & 41.67 & 33.33 & 33.33 & 36.11 \\
        \hspace{1em}TCD & 44.00 & 32.67 & \textbf{35.00} & 37.22 \\
        \hspace{1em}TA-AE & 37.67 & 30.00 & 33.00 & 33.56 \\
        \hspace{1em}Ours & \textbf{46.33} & \textbf{34.67} & 34.78 & \textbf{38.59} \\
        \bottomrule
    \end{tabular}
\end{table}

\begin{table}[h]
    \centering
    \caption{Ablation study and sensitivity analysis.}
    \label{tab:ablation_main}
    \begin{tabular}{llcc}
        \toprule
        \textbf{Category} & \textbf{Variant} & \textbf{STH} & \textbf{TSH} \\
        \midrule
        \multicolumn{2}{c}{Vanilla} & 60.55 & 59.17 \\
        \midrule
        \multirow{3}{*}{FDA Ablation} 
         & FDA only & 62.90 & 61.50 \\
         & FDA + Attn & 63.50 & 61.90 \\
         & Random frames & 61.50 & 60.30 \\
        \midrule
        \multirow{2}{*}{TDS Ablation} 
         & w/o FDA weights & 62.50 & 61.10 \\
         & TDS + FDA weights & 63.80 & 62.20 \\
        \midrule
        \multirow{5}{*}{Sensitivity $(\gamma_\mathcal{T}, \gamma_\mathcal{P})$} 
         & (1.0, 1.0) & 60.55 & 59.17 \\
         & (1.2, 1.0) & 63.30 & 61.70 \\
         & (1.0, 0.5) & 61.50 & 59.80 \\
         & (1.2, 0.5) & \textbf{67.40} & \textbf{68.22} \\
         & (1.5, 0.5) & 66.90 & 67.50 \\
        \bottomrule
    \end{tabular}
\end{table}

\subsection{Ablation Study}
\label{sec:ablation}

To evaluate the effectiveness of each component, we conduct ablation studies on VidHalluc using Qwen2.5-VL-7B. Table~\ref{tab:ablation_main} reports the main results on frame selection, subspace construction, and factor sensitivity. Additional analyses of fusion weights, subspace ranks, layer choices, and decomposition designs are provided in the supplementary material.

\textbf{Frame Distinctiveness Analysis.} The distinctiveness score $\phi_t$ measures how differently a frame relates to others in middle layers. High-scoring frames mark scene transitions or salient temporal moments. As shown in Table~\ref{tab:ablation_main}, FDA alone improves STH and TSH by 2.35 and 2.33 points over the vanilla model, confirming that internal discriminative signals encode underutilized temporal information. Cross-attention refinement yields gains of 0.60 and 0.40 points, validating the benefit of aligning FDA weights with the LLM's cross-modal reasoning pathway. FDA also outperforms random frame selection by 2.00 points on STH and 1.60 points on TSH, confirming that hidden-state geometry captures temporally informative boundaries more effectively than uniform sampling.

\begin{figure*}[t]
  \centering
  \includegraphics[width=\textwidth]{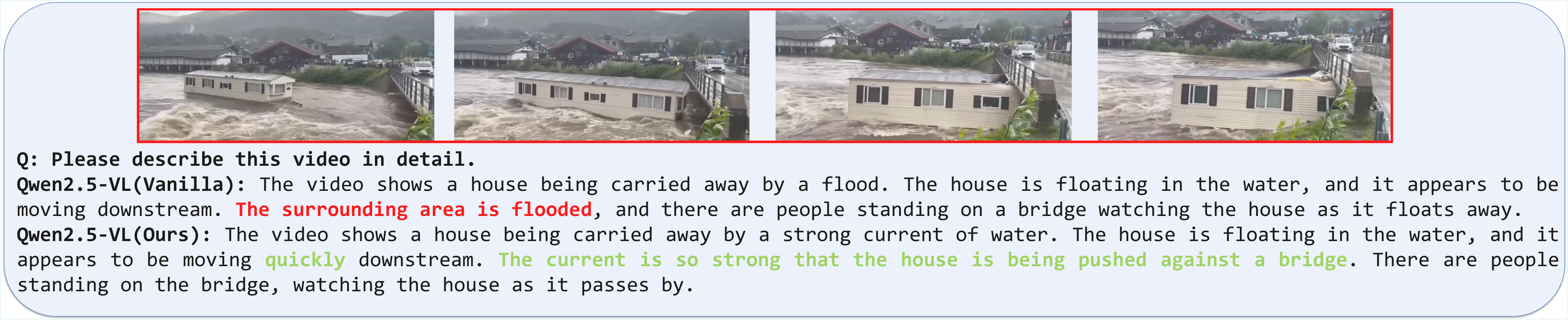}
  \caption{Qualitative example on EventHallusion with Qwen2.5-VL-7B, where bold green text denotes correct descriptions with accurate temporal details, and bold red text denotes hallucinated descriptions.}
  \label{fig:case_study}
\end{figure*}

\begin{figure}[t]
  \centering
  \includegraphics[width=0.4\textwidth]{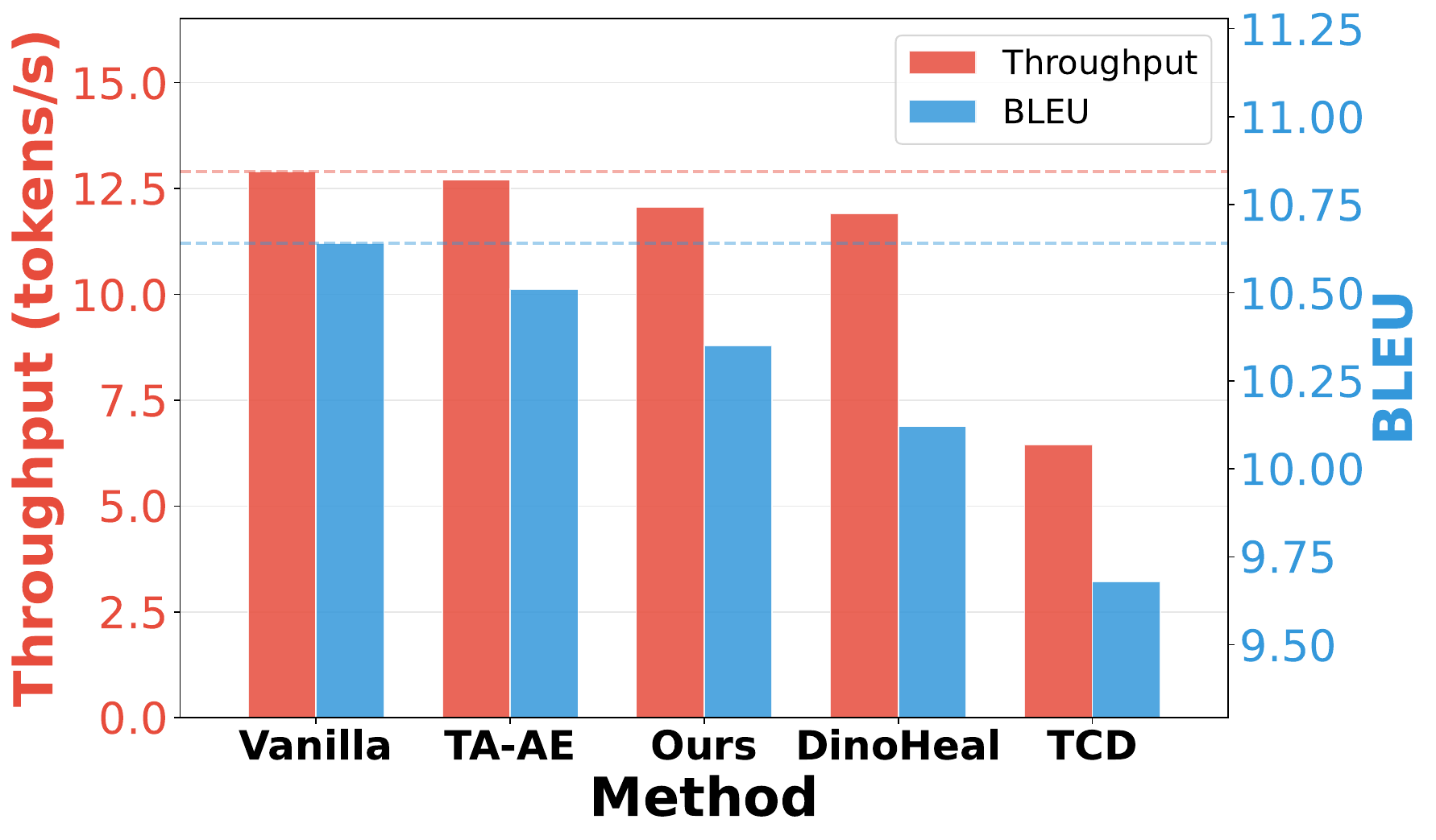}
    \caption{Joint comparison of inference throughput and BLEU scores across different methods.}
  \label{fig:comparison_bar_chart}
\end{figure}

\textbf{Temporal Delta Subspace.} Constructing the subspace from inter-frame differences is critical. As shown in Table~\ref{tab:ablation_main}, removing FDA weights reduces STH and TSH by 1.30 and 1.10 points relative to TDS with FDA weights. The consistent degradation across both metrics indicates that salient-frame weighting improves temporal subspace construction rather than favoring a single hallucination type. This suggests that temporally salient frames provide more informative and discriminative difference vectors, validating the benefit of frame-level saliency in TDS.

\textbf{Scaling Factor Sensitivity.} The enhancement mechanism independently rescales the temporal ($\mathcal{T}$) and prior ($\mathcal{P}$) components. As shown in Table~\ref{tab:ablation_main}, enhancing temporal evidence or suppressing prior information alone improves performance, while combining both yields the strongest results. The setting $(1.2, 0.5)$ achieves 67.40 STH and 68.22 TSH, indicating that the two operations address complementary temporal errors. Additional sensitivity results show that performance degrades smoothly around this setting.

\subsection{Efficiency and Linguistic Quality}

\textbf{Inference efficiency.} We evaluate inference throughput across all methods and diverse backbone models. As shown in Figure~\ref{fig:comparison_bar_chart}, TCD suffers a severe throughput reduction, reaching only 6.45 tokens/s compared to 12.90 for Vanilla, which is expected given its multiple forward passes. SPOD achieves 12.07 tokens/s, surpassing both DINO-HEAL at 11.91 tokens/s and TCD, while remaining close to Vanilla. These results confirm that our method incurs only marginal overhead relative to an additional full forward pass.

\textbf{Linguistic quality.} We further assess answer consistency using BLEU. Results in Figure~\ref{fig:comparison_bar_chart} reveal that TCD incurs the largest degradation, dropping to 9.68 BLEU from Vanilla's 10.64, a decrease of 0.96 points. DINO-HEAL also declines to 10.12 (a 0.52-point loss). Conversely, SPOD achieves a relatively high BLEU score of 10.35 among all methods, with only a 0.29-point gap from Vanilla. These findings suggest that the contrasting process in TCD introduces unwanted noise into surface-level language generation, whereas orthogonal subspace rescaling in SPOD preserves lexical consistency.

\subsection{Case Study}

To illustrate how SPOD improves temporal reasoning, we examine qualitative examples from EventHallusion. Fig.~\ref{fig:case_study} shows a clip in which a mobile home drifts along a swollen river and is pinned against a bridge while onlookers watch, requiring fine-grained temporal grounding rather than reliance on a static flooding prior. Under the same prompt, the vanilla model produces a template-like flood description and misses the collision moment and preceding trajectory. With SPOD, FDA up-weights frames with pronounced hidden-state changes, while TDS amplifies temporal delta directions so that generation follows inter-frame evidence rather than coarse scene semantics. This behavior aligns with the event-grounding objective of EventHallusion. Similar patterns appear in VidHalluc transition clips, where vanilla models fixate on the opening scene, whereas FDA highlights boundary frames and TDS captures representational shifts. These examples demonstrate that SPOD shifts reliance from static visual-language priors toward temporal structure encoded in hidden-state dynamics, providing a pathway for grounding video understanding in temporal evidence.

\section{Conclusion}

In this paper, we present SPOD as a training-free approach for temporal hallucination mitigation in Video-LLMs. Our insight is that Video-LLMs encode temporal evidence in hidden representations but fail to exploit it during inference. Through systematic analysis, we find that important frames exhibit higher hidden-state variance and inter-frame differences form a low-dimensional subspace. Motivated by these findings, SPOD introduces FDA and TDS to recover suppressed temporal evidence in a single forward pass. Extensive experiments on VidHalluc and other benchmarks show that SPOD consistently outperforms existing methods across model scales while maintaining inference efficiency. This work provides a mechanistic perspective for internal representation engineering.

\begin{acks}
This work was supported by the Beijing Natural Science Foundation (JQ24019), the National Natural Science Foundation of China (No. 62576047), the CIPS-SMP-Zhipu Large Model Fund (No. CIPS-SMP20250313), and the Open Fund of the Key Laboratory for Civil Aviation Collaborative Air Traffic Management Technology and Applications (No. 2025-001).
\end{acks}

\bibliographystyle{ACM-Reference-Format}
\balance
\bibliography{references}

\end{document}